\documentclass[]{sig-alternate}

\usepackage{multicol,times,algorithm,algorithmic}

\usepackage{epsfig,amsmath}

\begin{document}


\pagestyle{empty}

\newcommand{\N}{\mathop{\rm I\kern-.2emN}}
\newcommand{\R}{\mathop{\rm I\kern-.2emR}}


%
\title{Deceptiveness and Neutrality\\ The ND Family of Fitness Landscapes}
\numberofauthors{1}

\author{
%
\alignauthor William Beaudoin, S\'{e}bastien Verel, Philippe Collard, Cathy Escazut \\
       \affaddr{University of Nice-Sophia Antipolis}\\
       \affaddr{I3S Laboratory}\\
       \affaddr{Sophia Antipolis, France}\\
       \email{\{beaudoin,verel,pc,escazut\}@i3s.unice.fr}
}

%
%


\maketitle


\begin{abstract}
When a considerable number of mutations have no effects on fitness values, the fitness landscape is said neutral.
In order to study the interplay between neutrality, which exists in many real-world applications, and performances of metaheuristics, it is useful to design landscapes which make it possible to tune precisely neutral degree distribution.
Even though many neutral landscape models have already been designed, none of them are general enough to create landscapes with specific neutral degree distributions.
We propose three steps to design such landscapes: first using an algorithm we construct a landscape whose distribution roughly fits the target one, then we use a simulated annealing heuristic to bring closer the two distributions and finally we affect fitness values to each neutral network.
Then using this new family of fitness landscapes we are able to highlight the interplay between deceptiveness and neutrality.
\end{abstract}


\category{I.2.8}{Artificial Intelligence}{Problem Solving, Control Methods, and Search.}



\terms{Algorithms, performance, design, experimentation.}

\keywords{Fitness landscapes, genetic algorithms, search, benchmark.}

\section{Introduction}

The Adaptative Landscape metaphor introduced by S.~Wright~ \cite{wright:rmicse} has dominated the view of adaptive evolution: 
an uphill walk of a population on a mountainous fitness landscape in which 
it can get stuck on suboptimal peaks.
Results from molecular evolution has changed this picture: Kimura's model~\cite{KIM:83} assumes that the overwhelming majority of mutations are 
either effectively neutral or lethal and in the latter case purged by 
negative selection. This assumption is called the neutral hypothesis. 
Under this hypothesis, dynamics of populations evolving on such 
neutral landscapes are different from those on adaptive landscapes: 
they are characterized by long periods of fitness stasis (population stated on 
a 'neutral network') punctuated by shorter periods of innovation with rapid fitness increases \cite{eldgould72}.
In the field of evolutionary computation, neutrality plays an important role in real-world problems: in design of digital circuits 
\cite{harvey96through} \cite{vassilev00advantages} \cite{Yu2002}, in evolutionary robotics \cite{Smith_CEC2002} \cite{seys04}. 
In those problems, neutrality is implicitly embedded in the genotype to phenotype mapping.

\subsection{Neutrality}

We recall a few fundamental concepts about fitness landscapes and neutrality (see
\cite{stadler-02} for a more detailed treatment).
A landscape is a triplet $(S, V, f)$ where $S$ is a set of
\textit{potential solutions} i.e. a search space,
 $V: S \rightarrow 2^{S}$, a \textit{neighbourhood} structure, is a function that
 assigns to every $s \in S$ a set of neighbours $V(s)$,
and $f:S \rightarrow \R$ 
is a fitness function that can be pictured as the ``height'' of the corresponding potential solutions. The neighbourhood is often defined by an operator like bitflip mutation.
A \textit{neutral neighbour} of $s$ is a neighbour with the same fitness $f(s)$.
The \textit{neutral degree} of a solution is the number of its neutral neighbours.
A fitness landscape is neutral if there are many solutions with high neutral degree.
A \textit{neutral network}, denoted $NN$, is a connected graph where vertices are solutions with the same fitness value and
two vertices are connected if they are neutral neighbours.

\subsection{Fitness Landscapes with Neutrality}

In order to study the relationship between neutrality, dynamics of Evolutionary Algorithms (EA) and search difficulty, 
some benchmarks of neutral landscapes have been proposed.
More often neutrality is either an add-on feature, as in NK-landscapes, or an incidental property, as in Royal-Road functions. In most cases the design acts upon the amount of solutions with the same fitness.
Royal-Road functions \cite{MIT-FOR-HOL:92} are defined on binary strings of length $N=n.k$ where $n$ is the number of blocks and $k$ the size of one block. 
The fitness function corresponds to the number of blocks which are set with $k$ bits value $1$ and neutrality increases with $k$.
Numerous landscapes are variant of NK-Landscapes \cite{kauffman93}.
The fitness function of an $NK$-landscape is a function $f: \lbrace 0, 1 \rbrace^{N} \rightarrow [0,1)$ defined on binary strings with $N$ bits.
An 'atom' with fixed epistasis level is represented by a fitness component $f_i: \lbrace 0, 1
\rbrace^{K+1} \rightarrow [0,1)$ associated to each bit $i$. 
It depends on the value at bit $i$ and also on the values at $K$ other epistatic bits. 
The fitness $f$ is the average of the values of the $N$ fitness components $f_i$.
Several variants of $NK$-landscapes try to reduce the number of fitness values in order to add some neutrality.
In $NKp$-landscapes \cite{barnett-1998}, $f_i(x)$ has a probability $p$ to be equal to $0$ ;
in $NKq$-landscapes \cite{newman98effect}, $f_i(x)$ is uniformly distributed in the interval $[0, q - 1] \cap \N$;
in \textit{Technological Landscapes} \cite{lobo04}, tuned by a natural number $M$, $f(x)$ is rounded so that it can only take $M$ different values.
For all those problems, neutrality is tuned by one parameter only: neutrality increases according to $p$ and decreases with $q$ or $M$ (see for example Figure \ref{fig_NKq}).

Dynamics of a population on a neutral network are complex, even on flat landscapes as shown by Derrida \cite{derrida91}.
The works \cite{bornberg99} \cite{NIM:99} \cite{nimwegen01} \cite{wilke01}, at the interplay of molecular evolution and optimization, 
study the convergence of a population on neutral networks. In the case of infinite population under mutation and selection, they show distribution on a $NN$ is only determined by the topology of this network. 
That is to say, the population converges to the solutions in the $NN$ with high neutral degree.
Thus, the neutral degree distribution is an important feature of neutral landscapes.

In order to study more precisely neutrality, for instance link between neutrality and search difficulty, we need for ``neutrality-driven design'' where neutrality really guides the design process.
In this paper we propose to generate a family of landscapes where it is possible to tune accurately the neutral degree of solutions.


\section {ND-Landscapes}

In this section, we first present an algorithm to create a landscape with a given \textit{neutral degree distribution}. Then we will refine the method to obtain more accurate landscapes and finally we will study time and space complexity of the algorithm.

\subsection{An algorithm to design \\ small ND-Landscape}

We now introduce a simple model of neutral landscapes called ND-Landscapes where N refers to the number of bits of a solution and D to the neutral degree distribution.
In this first step our aim is to provide an exhaustive definition of the landscape assigning one fitness value to each solution.
We fix N to 16 bits and so the size of search space is $2^{16}$.
Building a ND-Landscape is done by splitting the search space into neutral networks.
However the fitness value of each neutral network has no influence on the neutrality. This is why these fitness values are randomly chosen.

Let D be an array of size N+1 representing a neutral degree distribution.
N and D are given as inputs and the algorithm (see algorithm \ref{CrePays2}) returns a fitness function f from $\{0,1\}^N$ to $\R$ such that the neutral degree distribution of the fitness landscape is similar to D. For more simplicity, we chose to give a different fitness value to each neutral network. We define RouletteWheel(D) as a random variable whose density is given by distribution D. It is directly inspired from the genetic algorithm selection operator.
For example:
let $\Delta$ be the following distribution :\\
 $\Delta$[0]=0\ \ \ \ $\Delta$[1]=0.25\ \ \ \ $\Delta$[2]=0.5\ \ \ \ $\Delta$[3]=0.25. RouletteWheel($\Delta$) will return value \textbf{1} in 25 $\%$ of the time, \textbf{2} in 50 $\%$ of the time and \textbf{3} in 25 $\%$ of the time. Figure 1 
shows the neutral networks of an ideal ND-Landscape (size=$2^5$) for the distribution $\Delta$.


 \begin{figure}

   \label{idealND}
   \psfig{figure=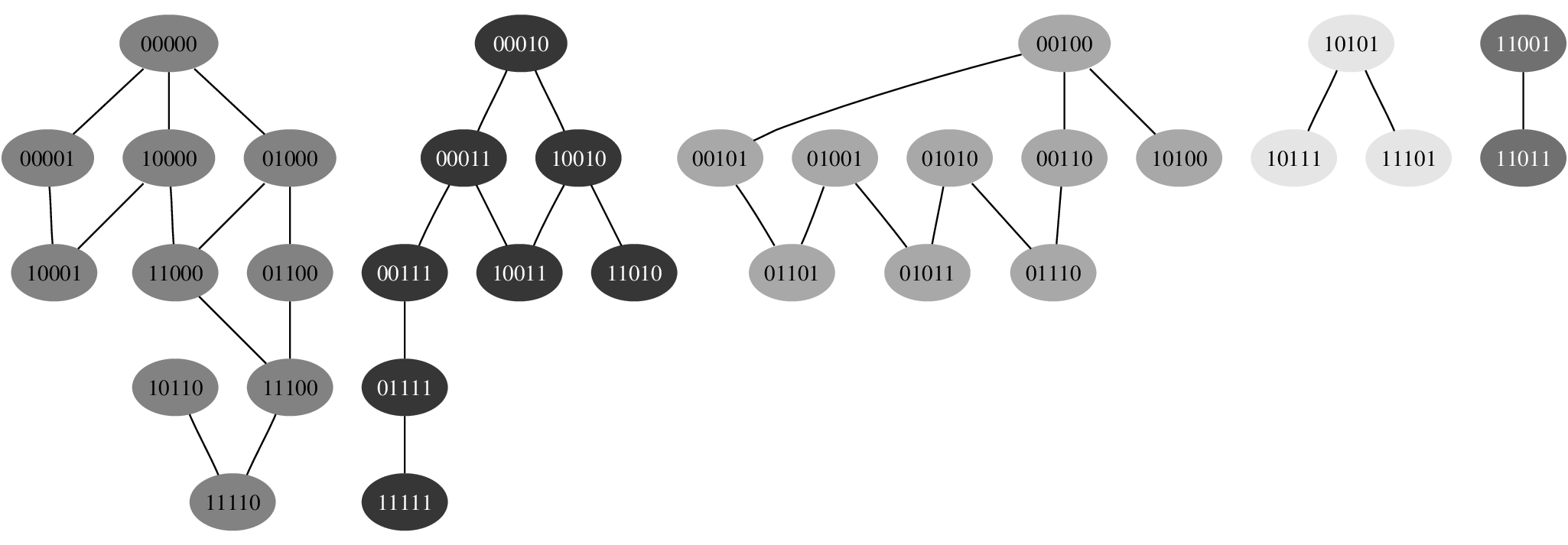,width=240pt,height=150pt}
   \caption{Example of a tiny ND-Landscape.
     Each node represents a solution and two nodes are connected if they have the same fitness value and are Hamming neighbours. In this example there are five neutral networks.}
 \end{figure}

\begin{figure*}[!tb] 
\begin{center}
\begin{tabular}{cc} 
\psfig{figure=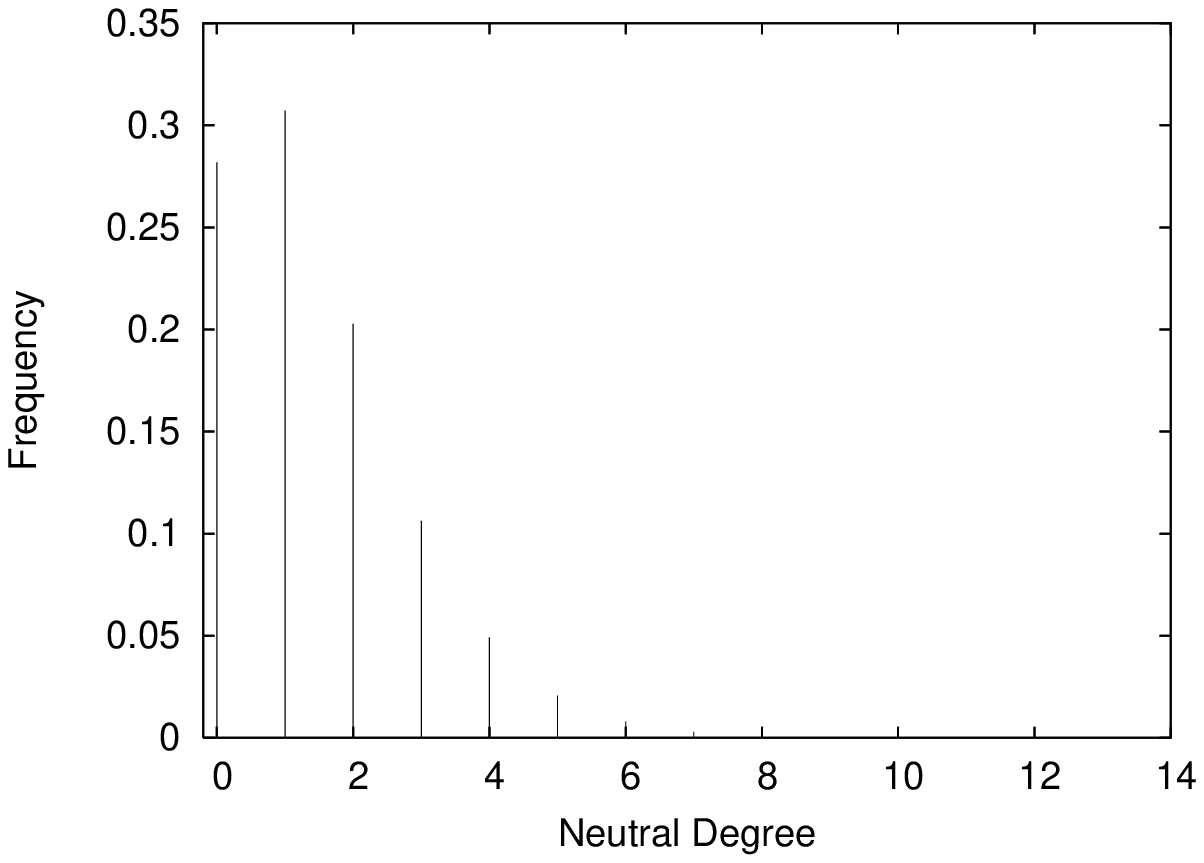,width=230pt,height=120pt} 
&
\psfig{figure=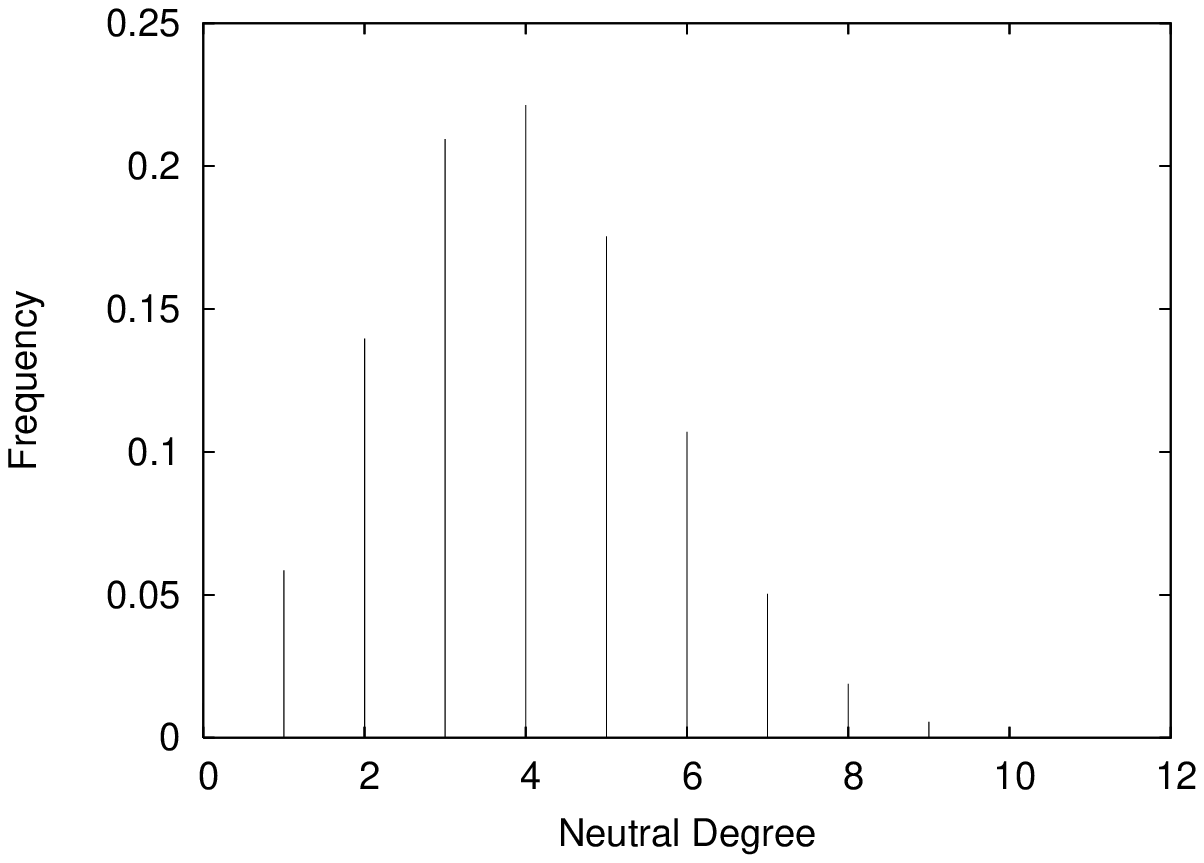,width=230pt,height=120pt}
\\
NKp with $N=16$, $K=5$, $p=0.8$ &
NKq with $N=16$, $K=4$, $q=2$
\\
average: 5.16  std dev:2.37 &
average: 3.94  std dev:1.74
\\

\psfig{figure=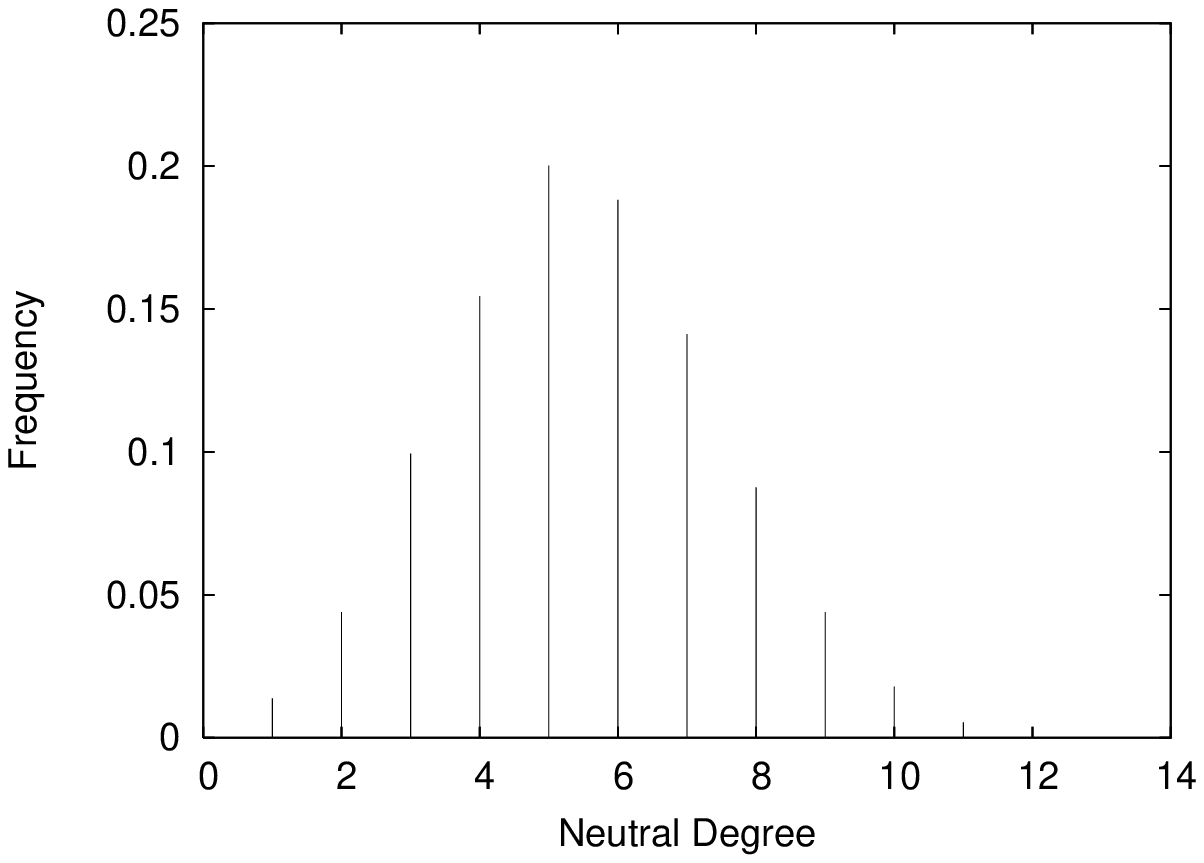,width=230pt,height=120pt}
&
\psfig{figure=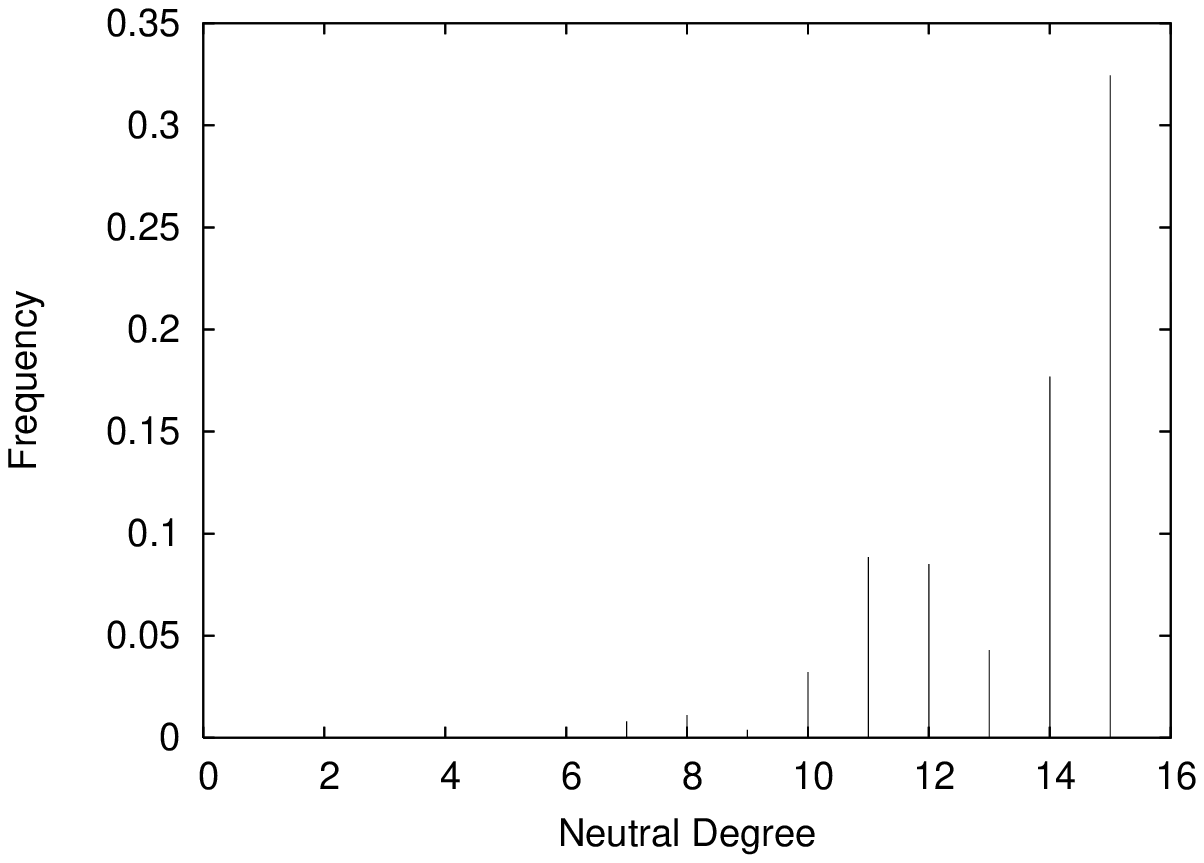,width=230pt,height=120pt} \\
Technological landscape with $N=16$, $K=4$, $M=20$
&
Royal Road with $N=16$, $n=4$, $k=4$ \\
average: 5.49  std dev:1.99 &
average: 14.0  std dev:2.00

\\
\end{tabular}
\end{center}
\caption{Neutral degree distribution for some neutral landscapes}
\label{fig_NKq}
\end{figure*}



 \begin{algorithm}
 \caption{Generation of ND-Landscapes}
 \label{CrePays2}

 \begin{algorithmic}
 \STATE $\forall{s \in S},\ $ f[s] $\leftarrow$ unaffected
 \STATE randomly choose one solution $s_{0}$.
 \STATE \textit{CandidatesList} $ \leftarrow$ S sorted by distance from $s_{0}$.
 \WHILE{not empty(\textit{CandidatesList}) }
 \STATE s $\leftarrow$ head(\textit{CandidatesList})

 \FOR{d = 0 to N}
        \STATE 
   \textbf{if} $s$ can't have \textit{d} neutral neighbours 
   \STATE \ \ \textbf{then} D'[d] $\leftarrow$ 0 
   \STATE \ \ \textbf{else} D'[d] $\leftarrow$ D[d] 
 \ENDFOR
 \STATE n $\leftarrow$ RouletteWheel(D'[d])
 \STATE Give a value to some unaffected neighbours so that \textit{s} has exactly n neutral neighbours and so that the neutral degrees of already chosen solutions ($\not\in$ \textit{CandidatesList}) are unchanged.
 \STATE D[n] $\leftarrow$ D[n] - $\frac{1}{2^N}$
 \STATE \textit{CandidatesList} $\leftarrow$ next(\textit{CandidatesList})
 \ENDWHILE        

\end{algorithmic}
\end{algorithm}

\subsection{A metaheuristic to improve the ND design}
Using algorithm \ref{CrePays2}, exhaustive fitness allocation does not create a landscape with a neutral degree distribution close enough to the input distribution. The reason is the fitness function is completely defined before the neutral degree of every solution has been considered. Hence, we use a simulated annealing metaheuristic 
 to improve the landscape created by algorithm \ref{CrePays2}. Here, simulated annealing is not used to find a good solution of a ND-Landscape but to adjust the landscape by modifying the fitness of some solutions such as
neutral distribution of a ND-Landscape be closer to the input distribution. The local operator is the changement of fitness value of one solution of the landscape, which can alter at most N+1 neutral degrees. The acceptance of a transition is determined by the difference between the distance to the input distribution before and after this transition.
The distance we use to compare two distributions is the root mean square :

$$ dist(D,D_0) = \sqrt{\sum_{i=0}^{N}(D[i]-D_0[i])^2} $$

\begin{figure*}[!tb] 
\begin{center}
\begin{tabular}{cc} 
\psfig{figure=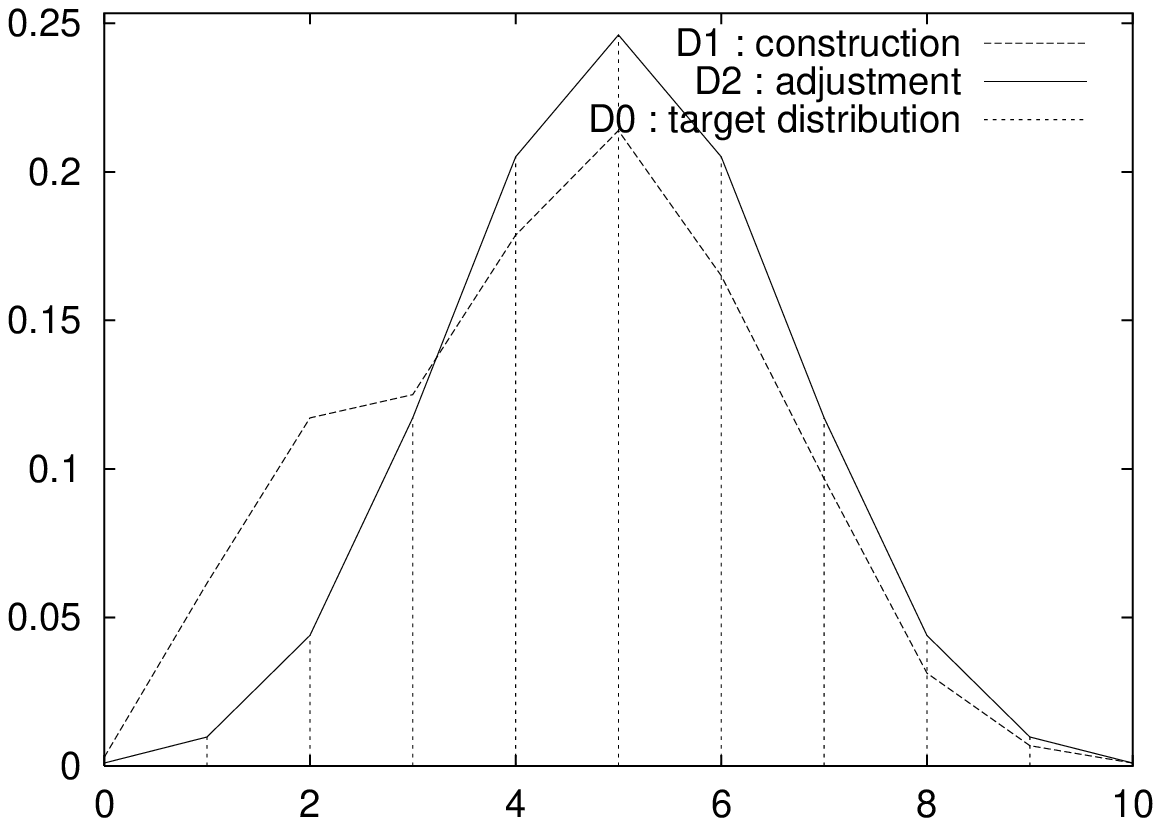,width=230pt,height=120pt} 
&
\psfig{figure=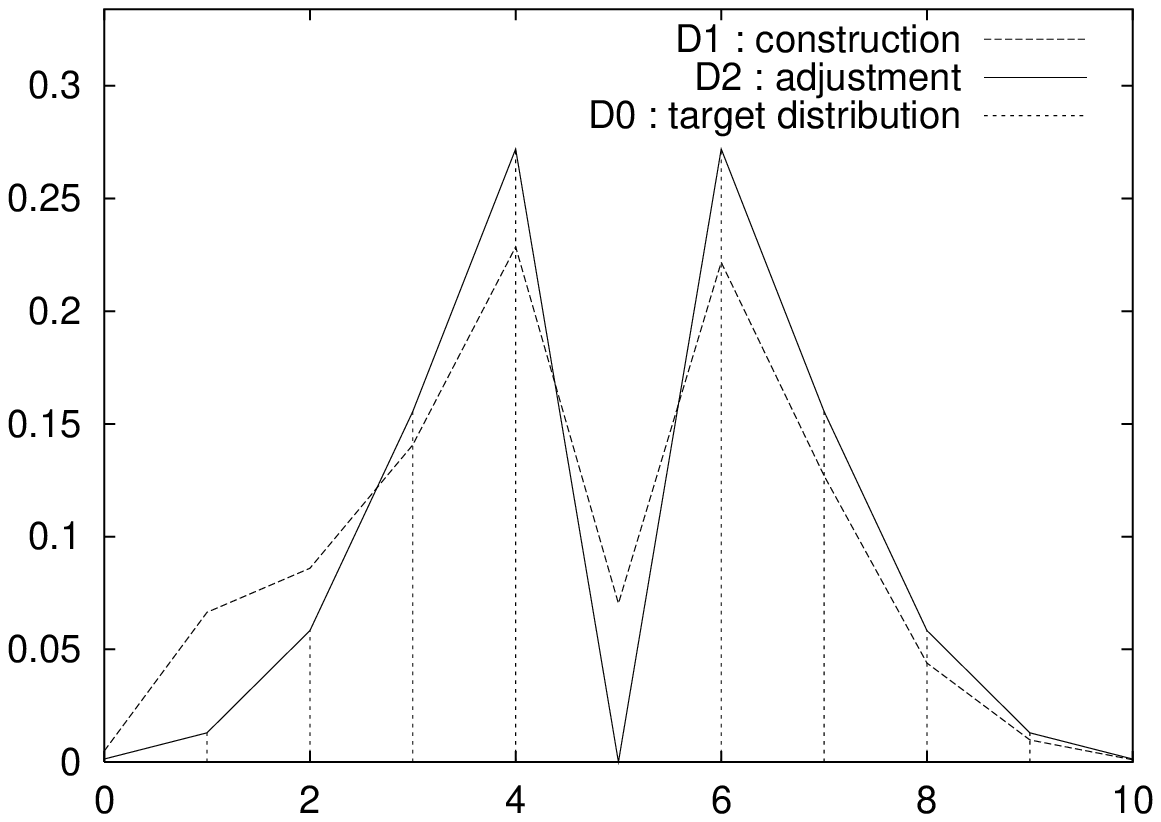,width=230pt,height=120pt}
\\
$dist(D_1,D_0) = 0.110 $ &
$dist(D_1,D_0) = 0.119 $
\\
$dist(D_2,D_0) = 0.00338 $ &
$dist(D_2,D_0) = 0.0175 $ 
\\
\psfig{figure=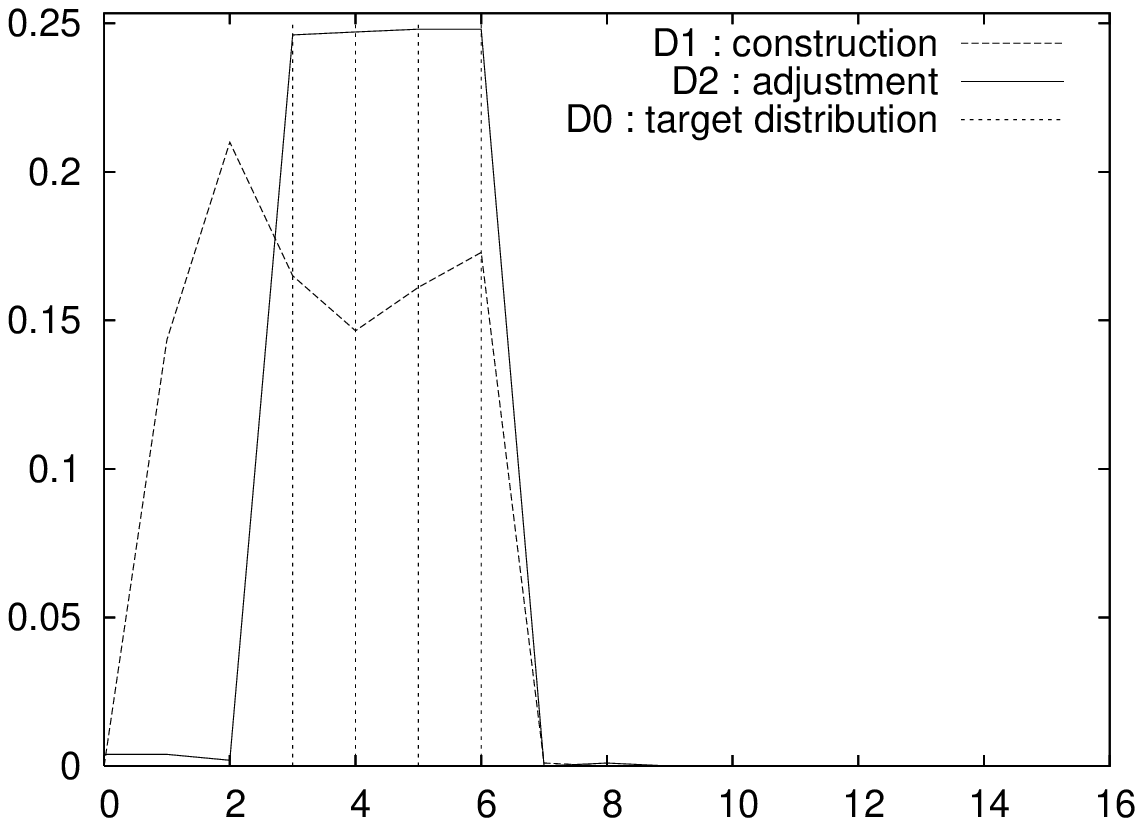,width=230pt,height=120pt}
&
\psfig{figure=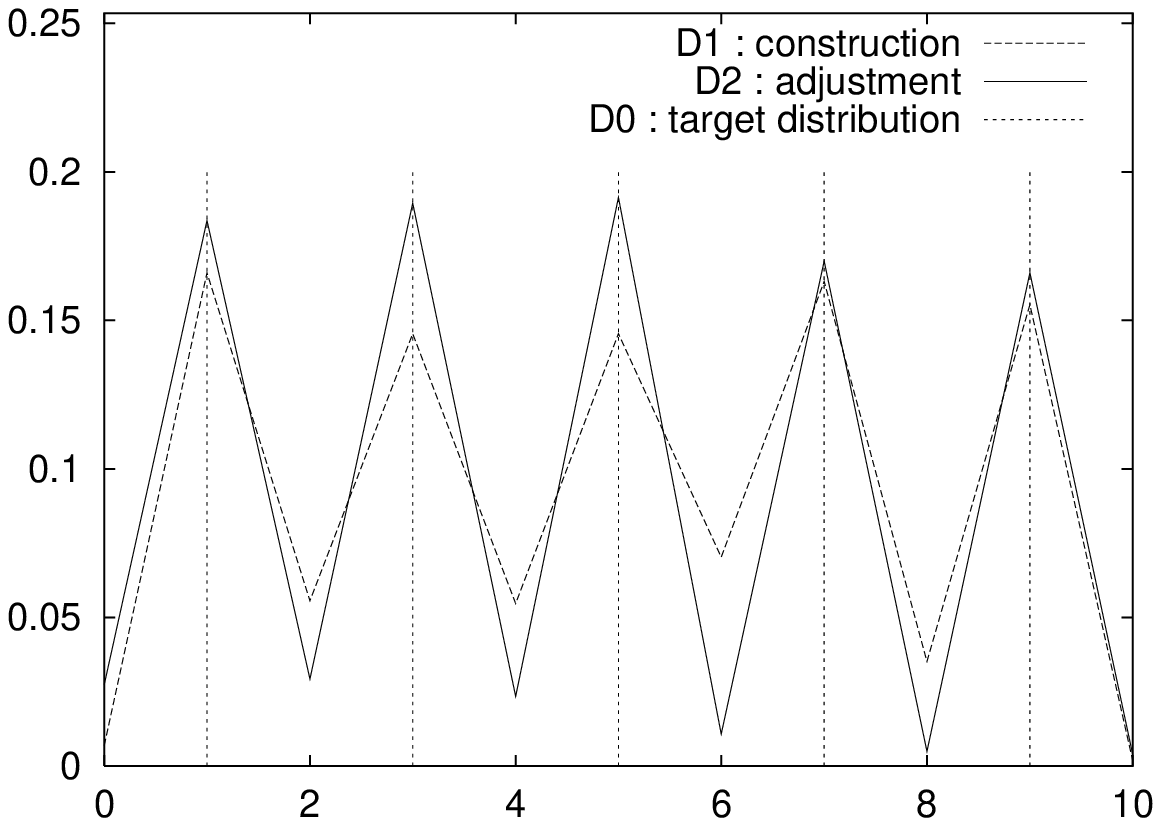,width=230pt,height=120pt} 
\\
$dist(D_1,D_0) = 0.0937 $ &
$dist(D_1,D_0) = 0.151 $
\\
$dist(D_2,D_0) = 0.00246$ &
$dist(D_2,D_0) = 0.0693 $ 
\\

\psfig{figure=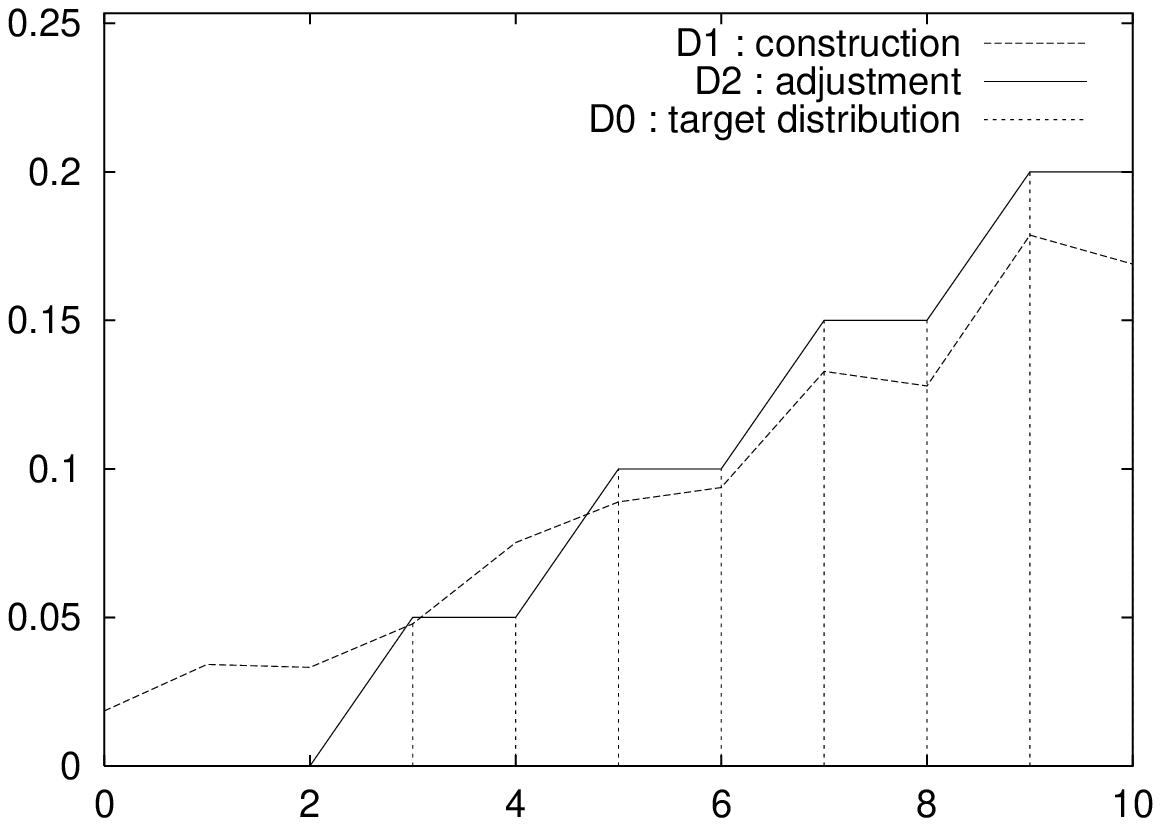,width=230pt,height=120pt}
&
\psfig{figure=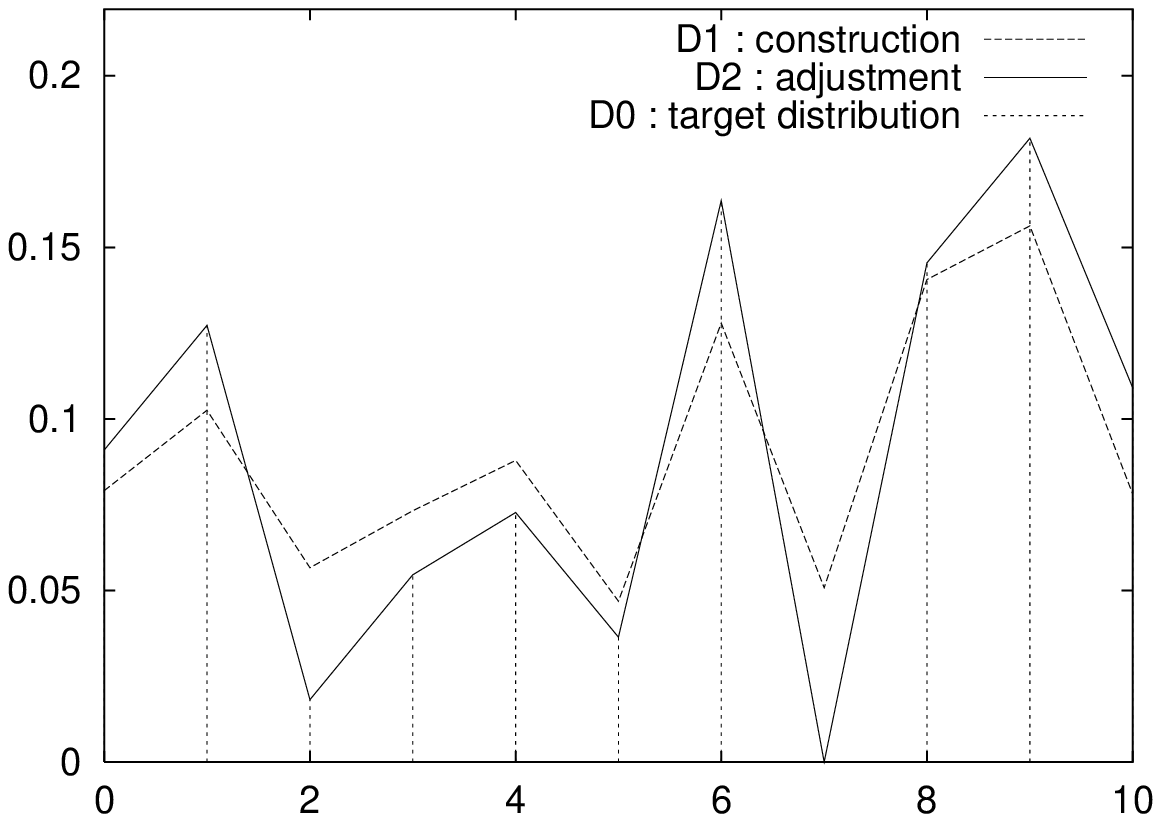,width=230pt,height=120pt} 
\\
$dist(D_1,D_0) = 0.0750 $ &
$dist(D_1,D_0) = 0.0917 $
\\
$dist(D_2,D_0) = 0.0103 $ &
$dist(D_2,D_0) = 0.00403 $ 
\\

\end{tabular}
\end{center}
\caption{Neutral Degree Distributions obtained by algorithm \ref{CrePays2} ($D_1$) and then adjusted by simulated annealing ($D_2$). Neutral degrees are on abcissa. Impulses represent the target distributions ($D_0$).}
\label{fig_DDN}
\end{figure*}

Simulated annealing appeared to be a fast and efficient method for this particular task.
Improvements made by simulated annealing are shown in figure \ref{fig_DDN}. Easiest neutral distributions to obtain seemed to be smooth ones like gaussian distributions (which are the most encountered when dealing with real or artificial neutral problems). On the other hand, sharp distributions (like the middle-right one of figure \ref{fig_DDN}) are really hard to approach. In addition, independently of the shape, distributions with higher average neutral degree are harder to approximate.


\subsection{Space and Time complexity}
  
  To create a landscape with a search space of size $2^N$, we use an array of size $2^N$ containing fitness values and a list of forbidden values for each solution. Thus we need  a memory space of size $\mathcal{O}(2^N \times N)$.
  Consequently the space complexity is : $\mathcal{O}(2^N \times N)$
  \par
In order to know what are the possible neutral degrees of an unaffected solution $s$, we must consider every interesting value for $s$ (the fitnesses of all neighbour solutions and a random value), and for each of these values, we must find out all possible neutral degrees. This can be done in a time $\mathcal{O}(N^2)$.
We evaluate the possible neutral degrees once for each solution.
Time allowed for simulated annealing is proportional to the time elapsed during construction.
Thus, the time complexity of the algorithm is $\mathcal{O}(2^NN^2)$.
\par
Consequently we can only construct ND-Landscapes with a small N ($\leq 16$) but we will see in section 4 how to create Additive Extended ND-Landscapes with far greater search spaces.






\subsection{Sizes of the generated Neutral Networks}

\begin{figure*}[!tb] 
\begin{center}
\begin{tabular}{cc}

\psfig{figure=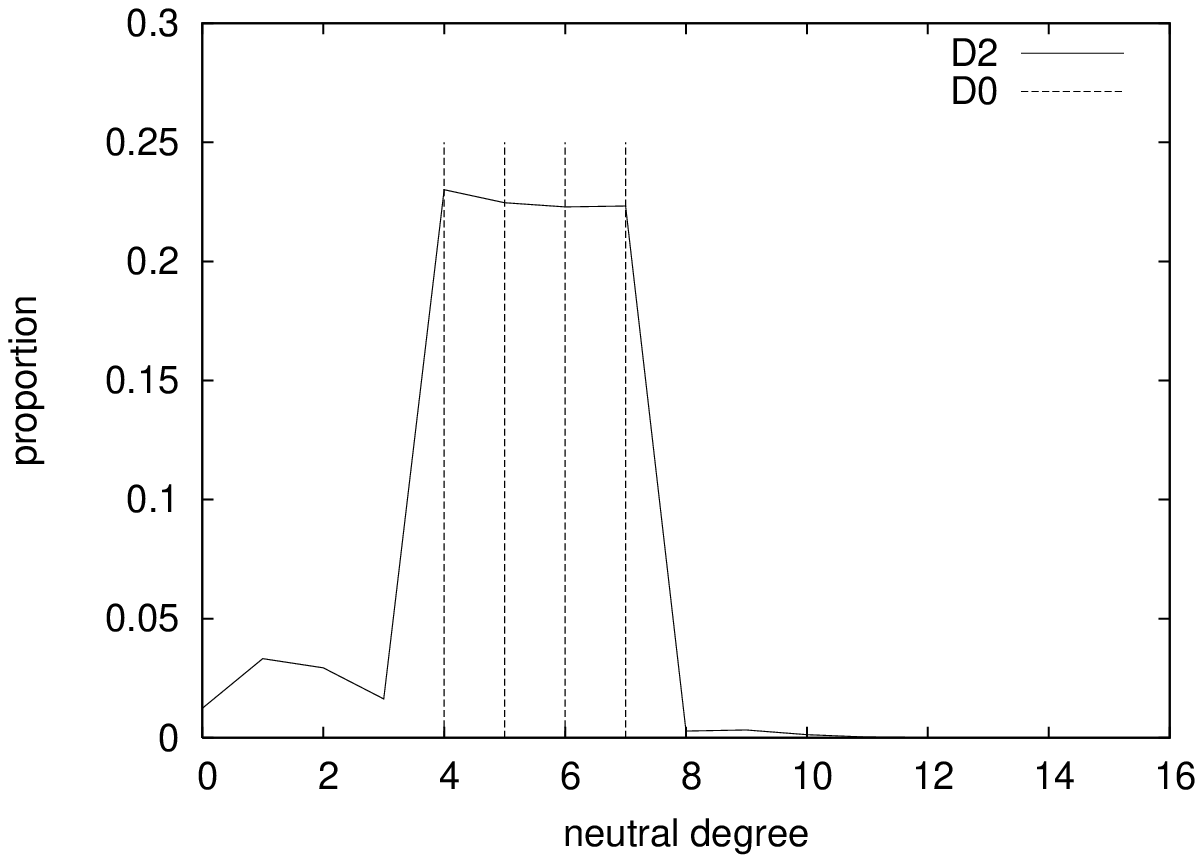,width=230pt,height=110pt} 
&
\psfig{figure=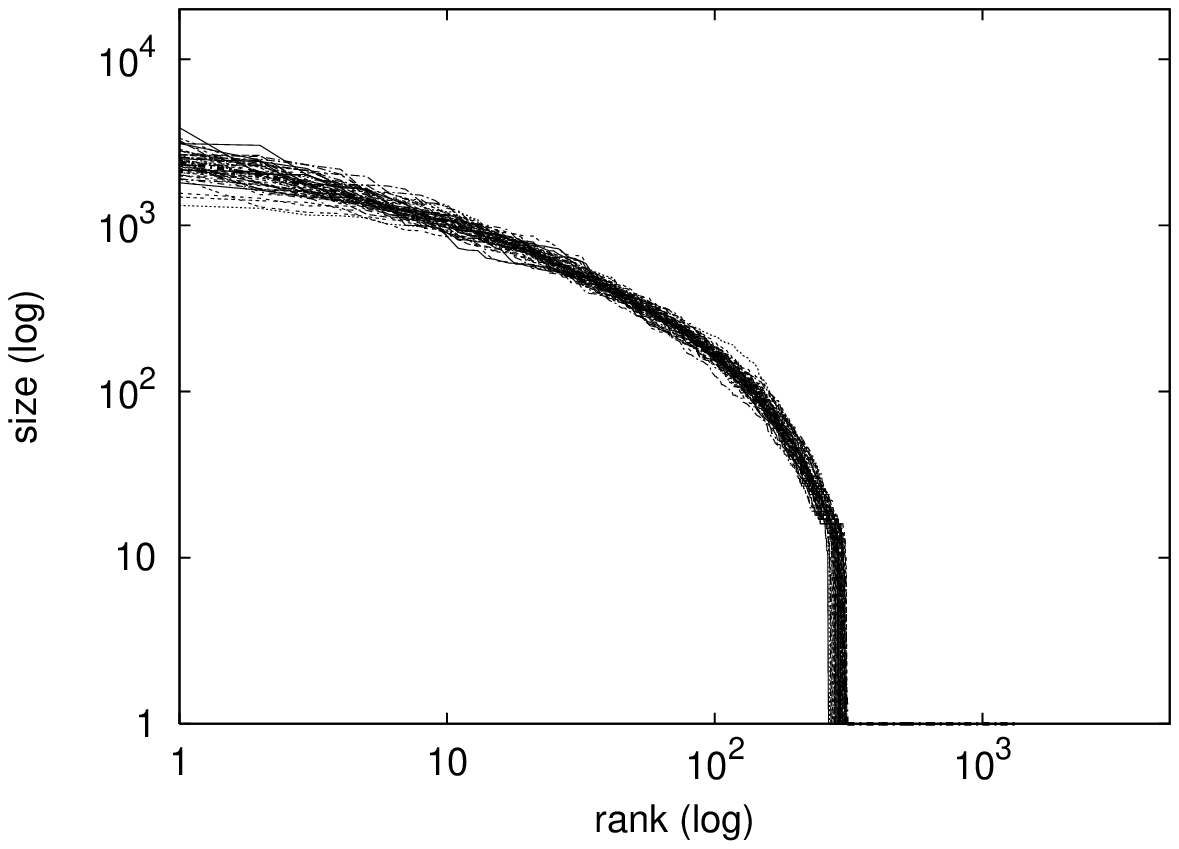,width=230pt,height=110pt}
\\
mean neutral degree 5.15 &
\ \ 
\\


\psfig{figure=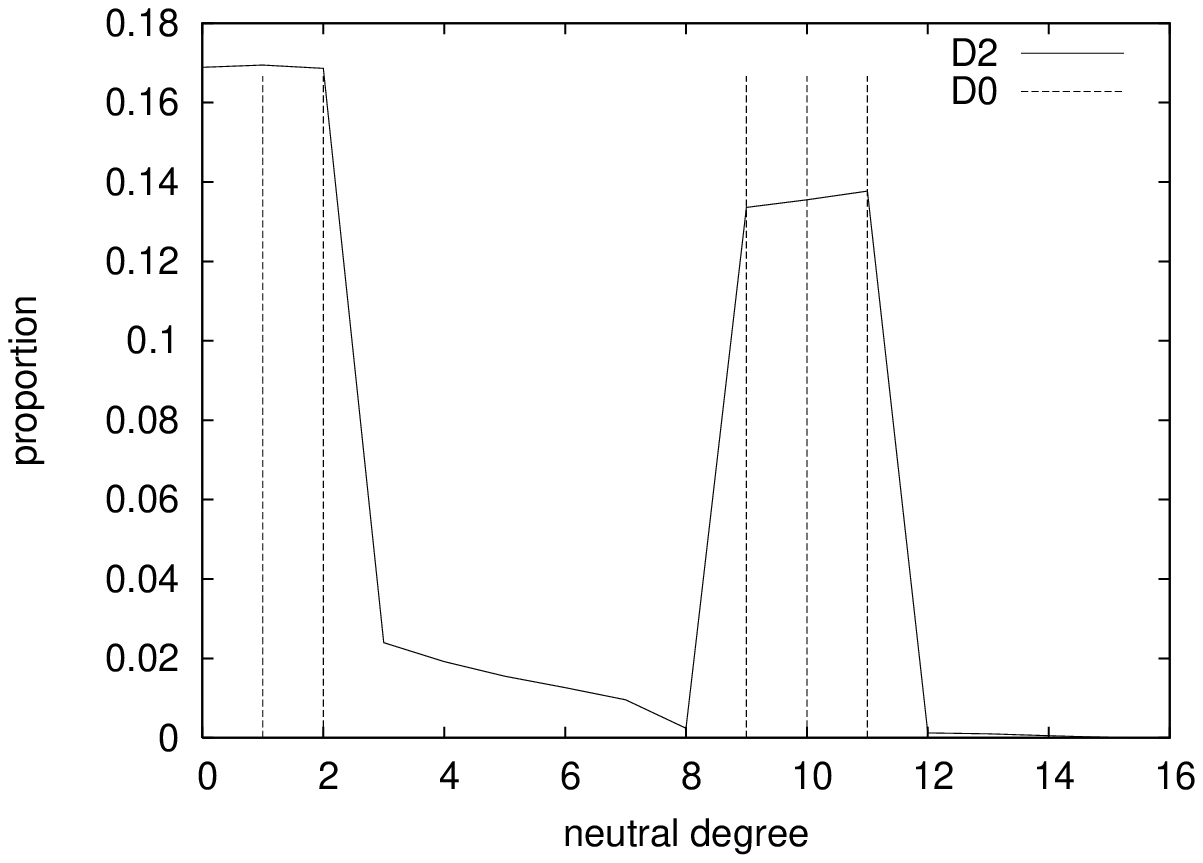,width=230pt,height=110pt} 
&
\psfig{figure=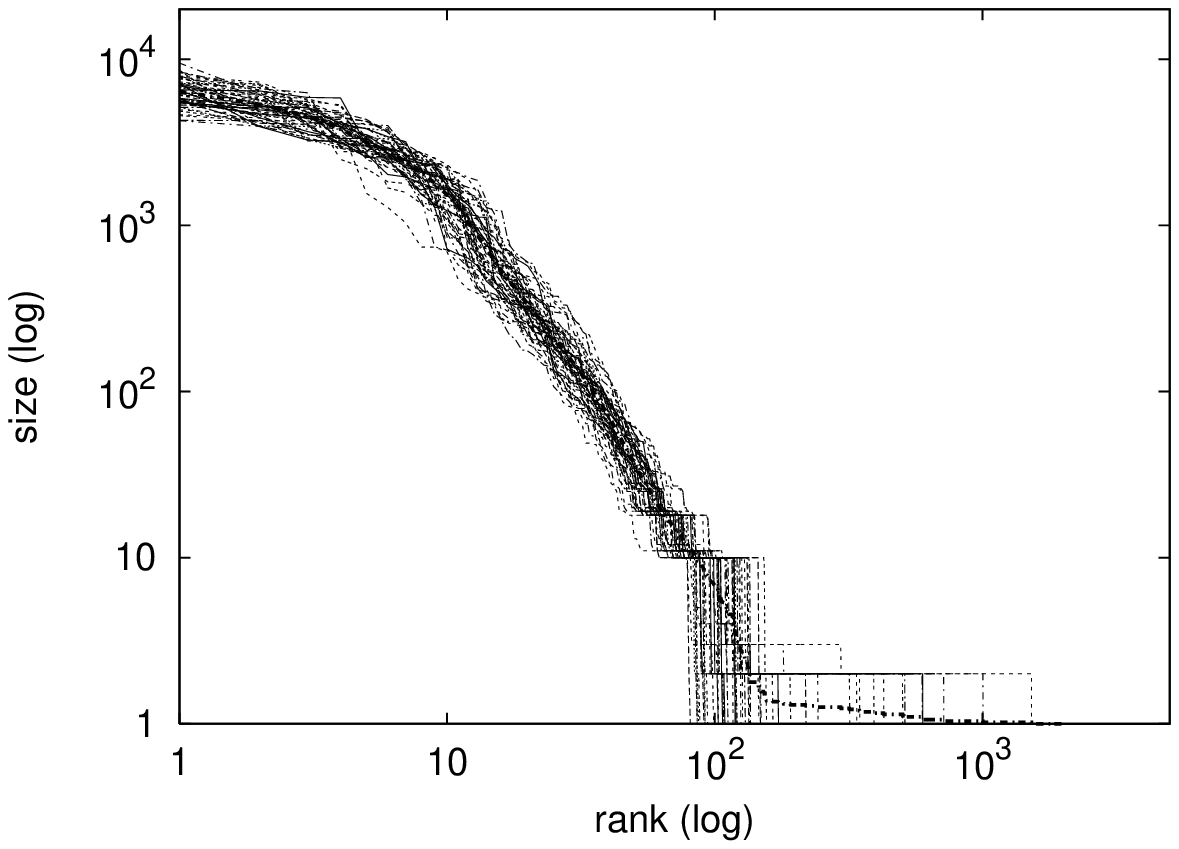,width=230pt,height=110pt}
\\
mean neutral degree 5.00 &
\ \ 
\\

\psfig{figure=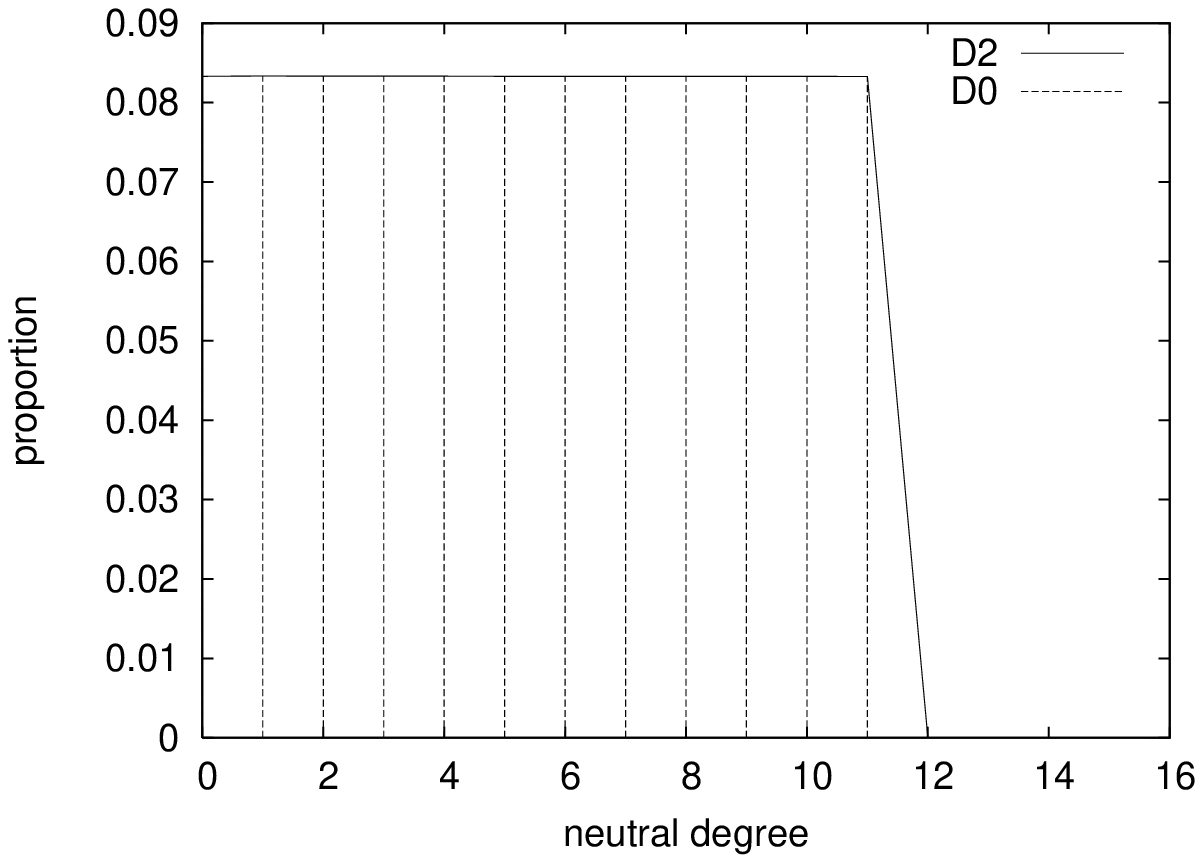,width=230pt,height=110pt} 
&
\psfig{figure=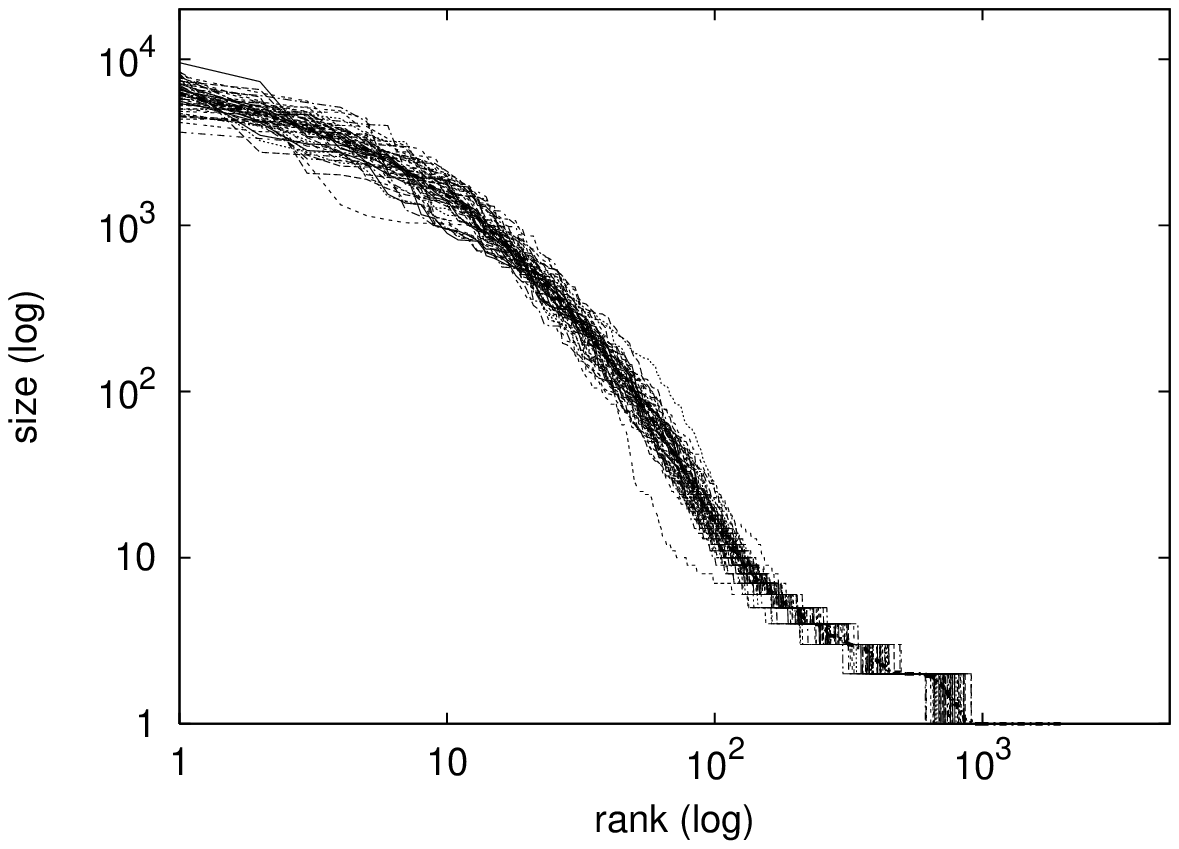,width=230pt,height=110pt}
\\
mean neutral degree 5.50 &
\ \ 
\\

\psfig{figure=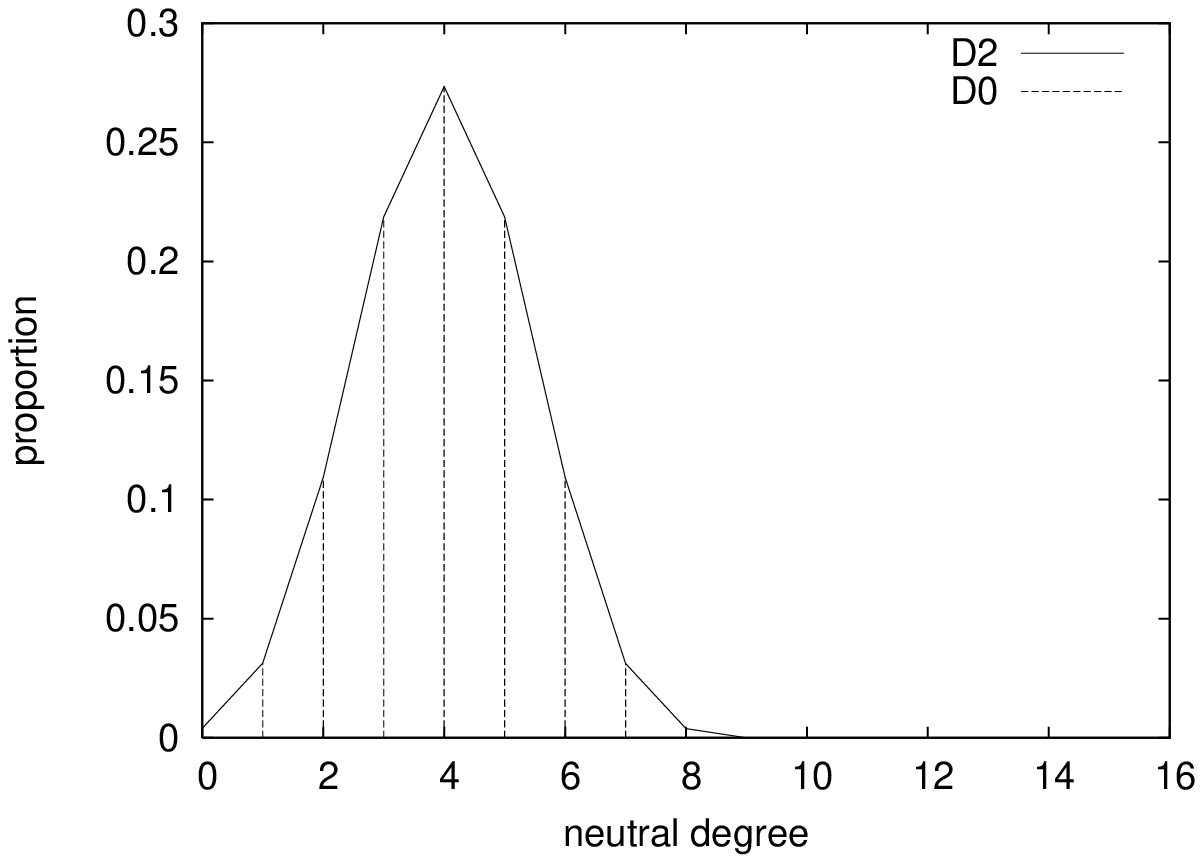,width=230pt,height=110pt} 
&
\psfig{figure=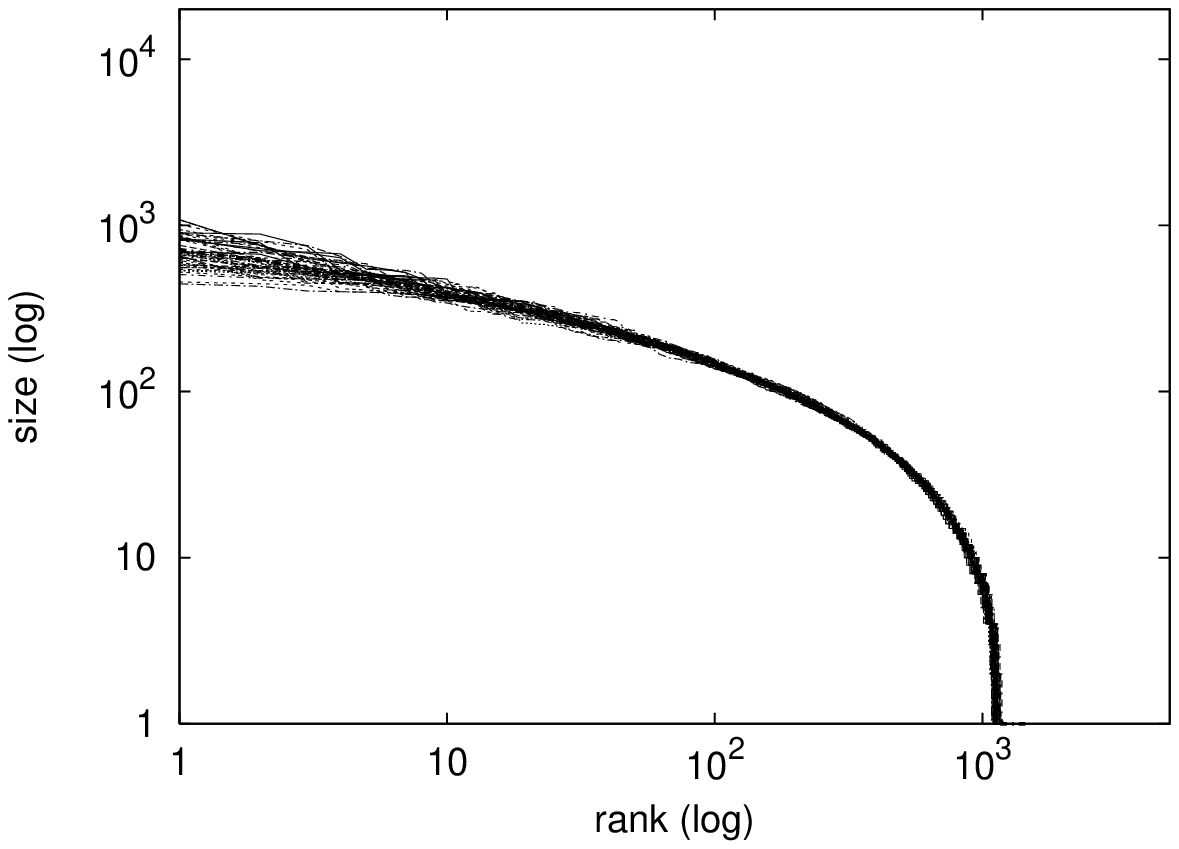,width=230pt,height=110pt}
\\
mean neutral degree 4.00 &
\ \ 
\\

\end{tabular}
\end{center}
\caption{Neutral networks sizes for various ND-landscapes}
\label{fig_rank}
\end{figure*}

Figure \ref{fig_rank} shows the diversity of sizes of neutral networks for 4 distributions.
For every distribution we created 50 different ND-Landscapes. 
Graphics on the left show the input and the mean resulting distribution.
Graphics on the right show all of the networks of these landscapes sorted by decreasing size with a logarithmic scale.
We clearly see that the neutral degree distribution is a really determining parameter for the structure of the generated landscape.

\section{Tuning deceptiveness \\ of ND-Landscapes}

Once we have generated a landscape with a specific neutral degree distribution, we can change the fitness value of all neutral networks without changing the neutral degree distribution (as long as we do not give the same fitness to two adjacent networks). Hence, for a given neutral distribution, we can tune the difficulty of a ND-Landscape.
For instance if each $NN$ have a random fitness value from $[0,1]$ then
the landscape is very hard to optimize.
Here, we will use the well known \textit{Trap} Functions \cite{deb93}
to affect fitnesses to $NN$ in order to obtain a ND-Landscape with
tunable deceptiveness.

\begin{figure}[!tb] 
\begin{center}

\begin{tabular}{cc} 
\psfig{figure=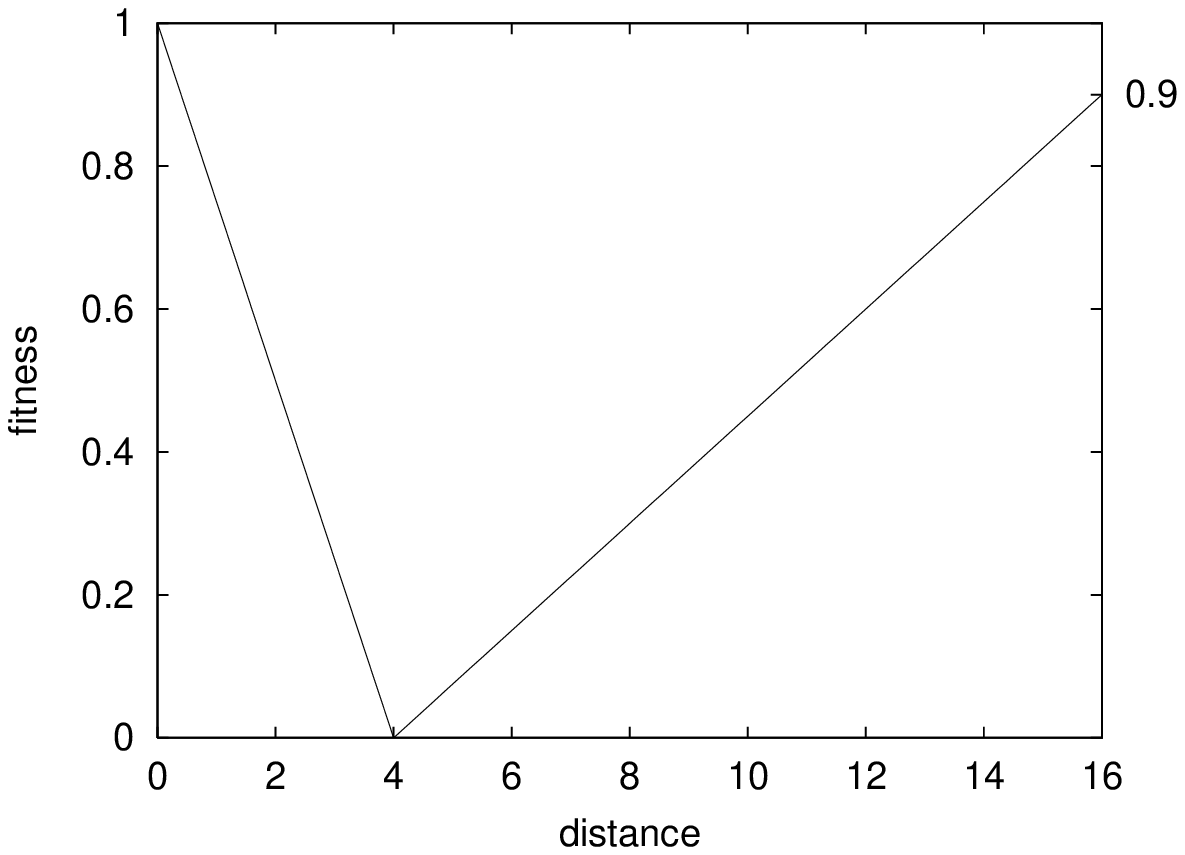,width=160pt,height=100pt} 
\\
(a) 
\\
\psfig{figure=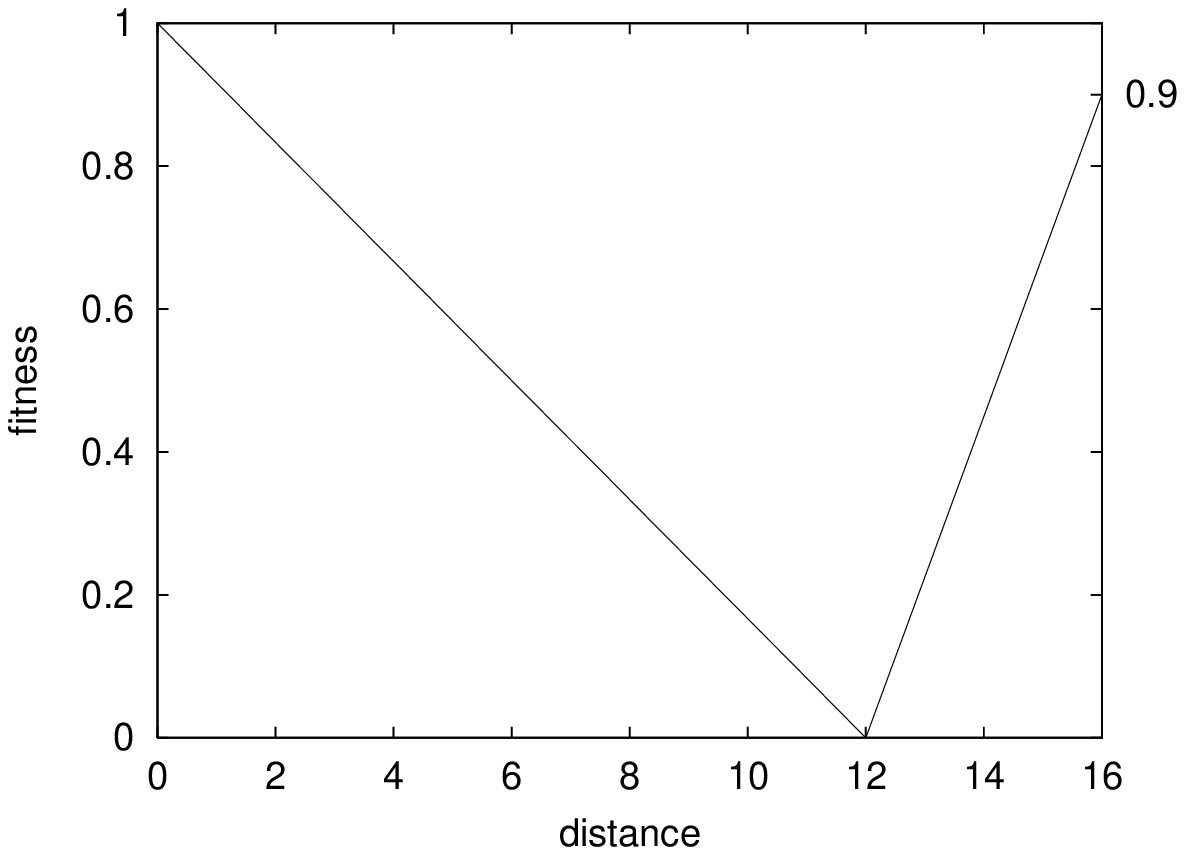,width=160pt,height=100pt}
\\
(b) 
\end{tabular}
\caption{Two trap functions:
Deceptive one with $b=0.25 $ and $ r=0.9$ (a),
Easy one with $b=0.75$ and $ r=0.9$ (b)
\label{traps}}
\end{center}
\end{figure}

The trap functions are defined from the distance to one particular solution. 
They admit two optima, a global one and a local one. 
They are parametrized by two values $b$ and $r$. 
The first one, $b$ allows to set the width of the attractive basin for
each optima, and $r$ sets theirs relative importance.
The function $f_T : \lbrace 0,1 \rbrace^N \rightarrow \R $ is so defined by:
\begin{equation*}
   f_{T}(x)= \begin{cases}
1.0 - \frac{d(x)}{b} & \text{if } d(x) \leq b,\\
\frac{r(d(x) - b)}{1.0 - b} & \text{elsewhere}
\end{cases}
\end{equation*}
where $d(x)$ is the Hamming distance to the global optimum, divided by $N$, between $x$ and
one particular solution.
The problem is most deceptive as $r$ is low and $b$ is high.
In our experiment, we will use two kinds of Trap functions with
$r=0.9$, one with $b=0.25$ and another one with $b=0.75$ (see figure
\ref{traps} (a) and (b)).

To affect a fitness value to each neutral network, we first choose the
optimum neutral network, denoted $NN_{opt}$, (for example the one containing the
solution $0^N$) and
set its fitness to the maximal value $1.0$.
Then, for each neutral network, we compute the distance $d$ between
its centroid\footnote{The centroid of a $NN$ is the string of the
frequency of appearance of bit value $1$ at each position.} and the
centroid of $NN_{opt}$ ;
finally the fitness value of the $NN$ is set according to a trap function\footnote{This trap function is defined for all real numbers between 0 and N}  
and the distance $d$.
In order to ensure that all adjacent networks have different fitness values,
it is possible to add a white noise to the fitness values of each $NN$.
In the following experiments, the length of bitstring is $N=16$. 
ND-landscapes are constructed with uniform neutral degree distributions.
We use the distributions defined by

\begin{equation*}
D_{p,w}[i] = \begin{cases}
 1/w & \text{if } i \in \{p, p+w-1\},\\
 0 & \text{elsewhere} \\
\end{cases}
\end{equation*}
where $p \in \lbrace 0,7 \rbrace$, $w \in \lbrace 3,4 \rbrace$,
and the two Trap functions defined in figure \ref{traps}.
For each distribution and each Trap function, $30$
landscapes were generated.

\subsection{Fitness Distance Correlation \\ of ND-Landscapes}
\label{sectionfdc}

To estimate the difficulty to search in these landscapes we will use a measure
introduced by Jones \cite{jones95b} called \textit{fitness distance
correlation} (FDC).
Given a set $F = \{f_1, f_2, ..., f_m\}$ of $m$ individual fitness values and
a corresponding set $D = \{d_1, d_2, ..., d_m\}$ of the $m$
distances to the global optimum, FDC is defined as:
$$FDC = \frac{C_{FD}}{\sigma_F \sigma_D}$$
where:
$$C_{FD} = \frac{1}{m} \sum_{i = 1}^{m} (f_i - \overline{f})
(d_i - \overline{d})$$ is the covariance of $F$ and $D$ and
$\sigma_F$, $\sigma_D$, $\overline{f}$ and $\overline{d}$ are the
standard deviations and averages of $F$ and $D$. 
Thus, by definition, FDC stands in the range $[-1,1]$. 
As we hope that fitness increases as distance to global optimum decreases, 
we expect that, with an ideal fitness function, FDC will
assume the value of $-1$. 
According to Jones \cite{jones95b}, problems can be
classified in three classes, depending on the value of the FDC coefficient:

\begin{itemize}

\item \textit{Deceptive problems} ($FDC \ge 0.15$), in which fitness increases
with distance to optimum.

\item \textit{Hard problems} ($-0.15 < FDC < 0.15$) in which there is
no correlation between fitness and distance.

\item \textit{Easy problems} ($FDC \le -0.15$) in which fitness increases
as the global optimum approaches.

\end{itemize}

Hard problems are in fact hard to predict, since in this case, the FDC
brings little information.
The threshold interval $[-0.15,0.15]$ has been empirically determined by Jones. 
When FDC does not give a clear indication i.e.,
in the interval $[-0.15,0.15]$, examining the scatterplot of fitness
versus distance can be useful.
The FDC has been criticized on the grounds that counterexamples can
be constructed for which the measure gives wrong results
\cite{alt97,quick98,Clergue:2002:GFBbCoTF}.

\begin{figure}[!tb]
\psfig{figure=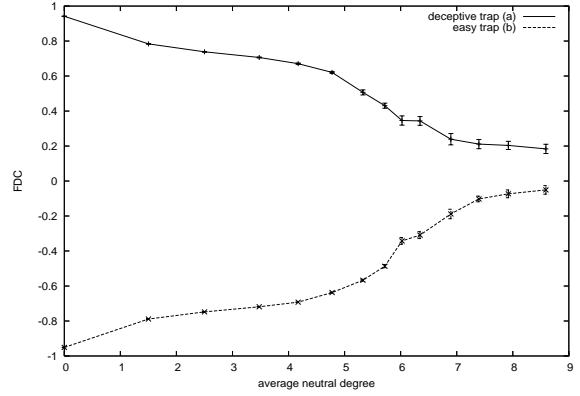,width=220pt,height=150pt} 
\caption{Average and standard deviation of FDC as a function of average neutral degree
for ND-landscapes created from easy and deceptive Trap functions.}
\label{fdcs}
\end{figure}

Figure \ref{fdcs} shows the average and standard deviation of FDC over
ND-Landscapes for each set of parameters, neutral distribution and deceptiveness.
The absolute value of FDC decreases as we generate ND-Landscapes with
more and more neutrality.
When adding neutrality, the landscapes are increasingly
flatter and thus less easy or deceptive. So neutrality smoothes correlation.
Adding neutrality to a deceptive landscape makes it easier and
adding neutrality to a easy landscape makes it harder.



\begin{figure}
\begin{tabular}{c}
\psfig{figure=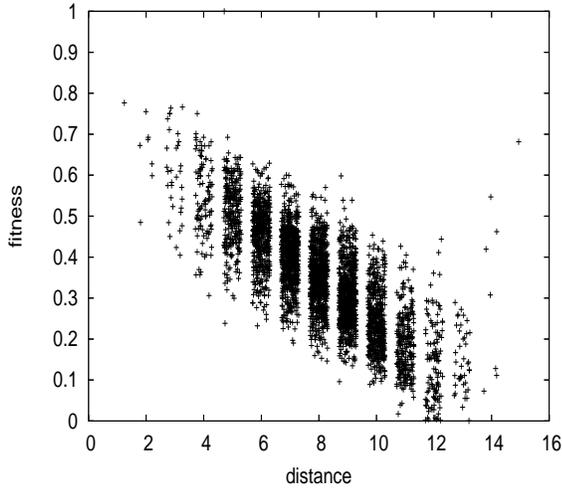,width=220pt,height=190pt} \\
(a) \\
\psfig{figure=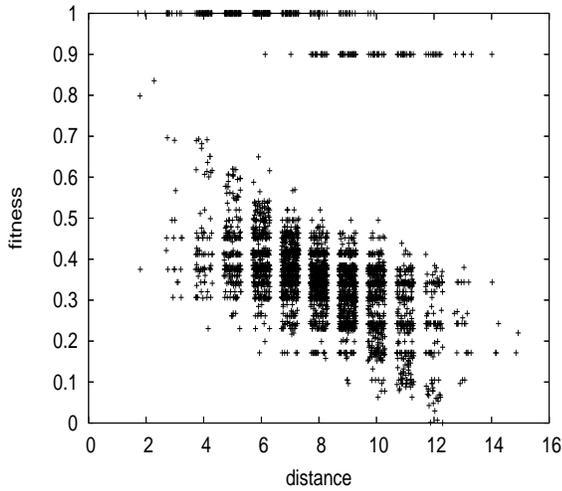,width=220pt,height=190pt} \\
(b) \\
\psfig{figure=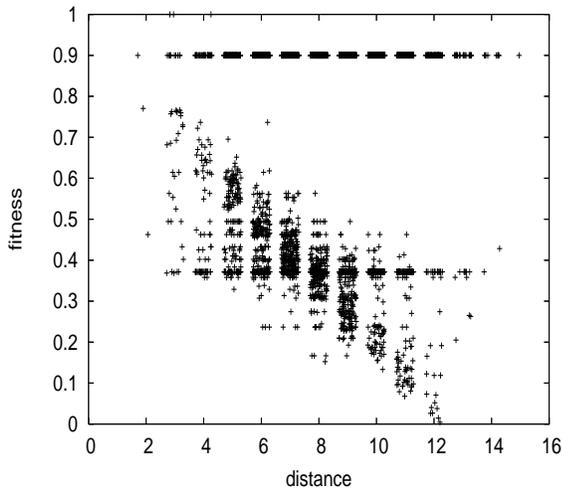,width=220pt,height=190pt} \\
(c) \\
\end{tabular}
\caption{ND-Landscapes : FDC scatter plot from easy trap (cf Figure \ref{traps} a)
with respective average neutral degree (a) 1.5  (b) 6  (c)  8.5}
\label{fdcc}
\end{figure}

\subsection{Genetic Algorithm Performances on ND-Landscapes}
\begin{figure}[!tb]
\psfig{figure=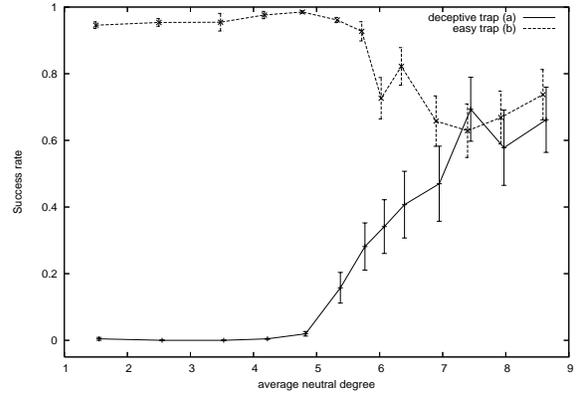,width=220pt,height=150pt} 
\caption{Average with standard deviation of success rate of a GA on ND-landscapes created from easy and deceptive trap functions}
\label{ag}
\end{figure}

In this section, difficulty is measured by genetic algorithm performances which is the success rate over $10³$ independent runs. In order to minimize the influence of the random creation of ND-Landscapes, we consider 30 different landscapes for each distribution $D$ and each trap function. For the GA, one-bit mutation and one-point crossover are used with rates of respectively 0.8 and 0.2. The evolution, without elitism and with 3-tournament selection of a population of 50 individuals took place during 50 generations.
Figure \ref{ag} shows average and standard deviation of GA performances over the 30 ND-landscapes.
Until the average neutral degree 5, the landscapes are fully deceptive (a) or fully easy (b).
Between neutral degree 5 and 7, the deceptiveness and the easiness decrease.
After neutral degree 7, the two trap functions have nearly a same good success rate (0.7). Whereas the FDC coefficient gave no information about performances, examining the scatter plot fitness/distance allowed to predict these good performances (figure \ref{fdcc} (c)).
When neutrality increases on deceptive trap, performances increase whereas on easy trap they decrease.
These results confirm conclusions found in section \ref{sectionfdc} from FDC measures.

\section{Additive extended \\ ND-Landscapes}

\begin{figure}[!tb] 
\begin{center}
\begin{tabular}{cc} 
\psfig{figure=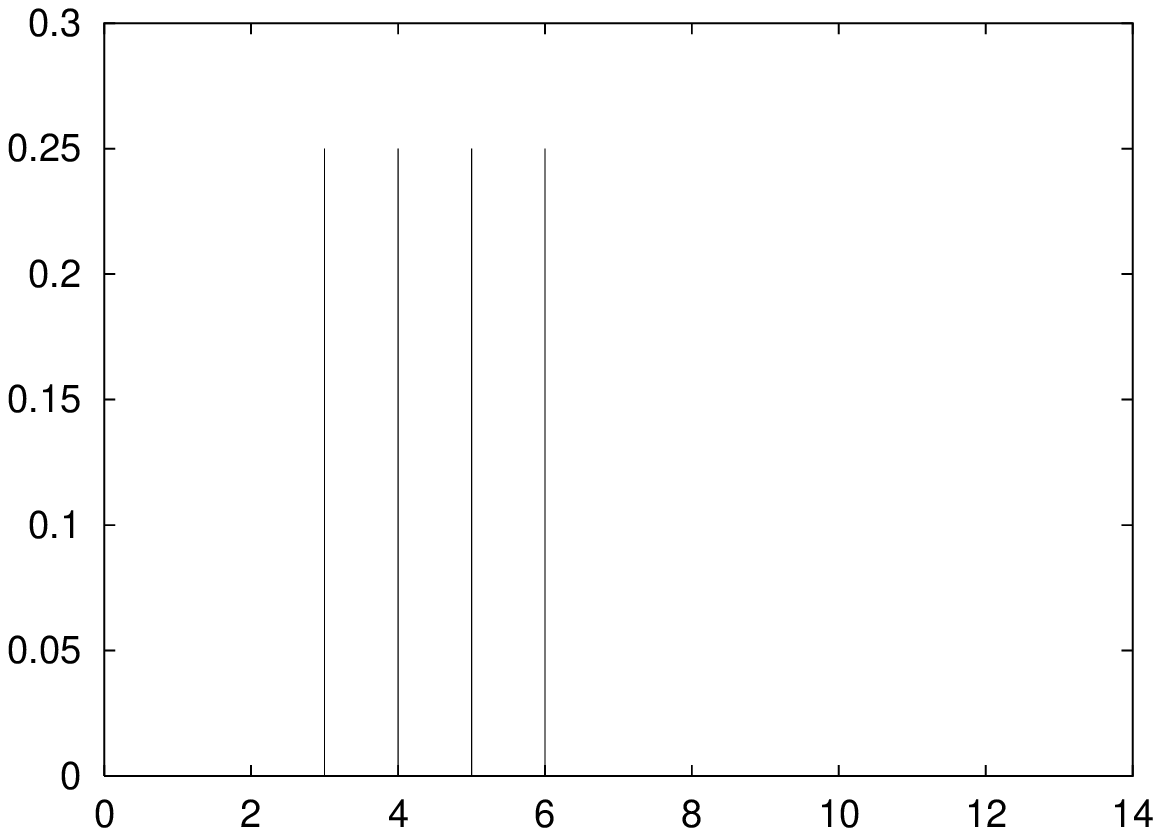,width=110pt,height=100pt} 
&
\psfig{figure=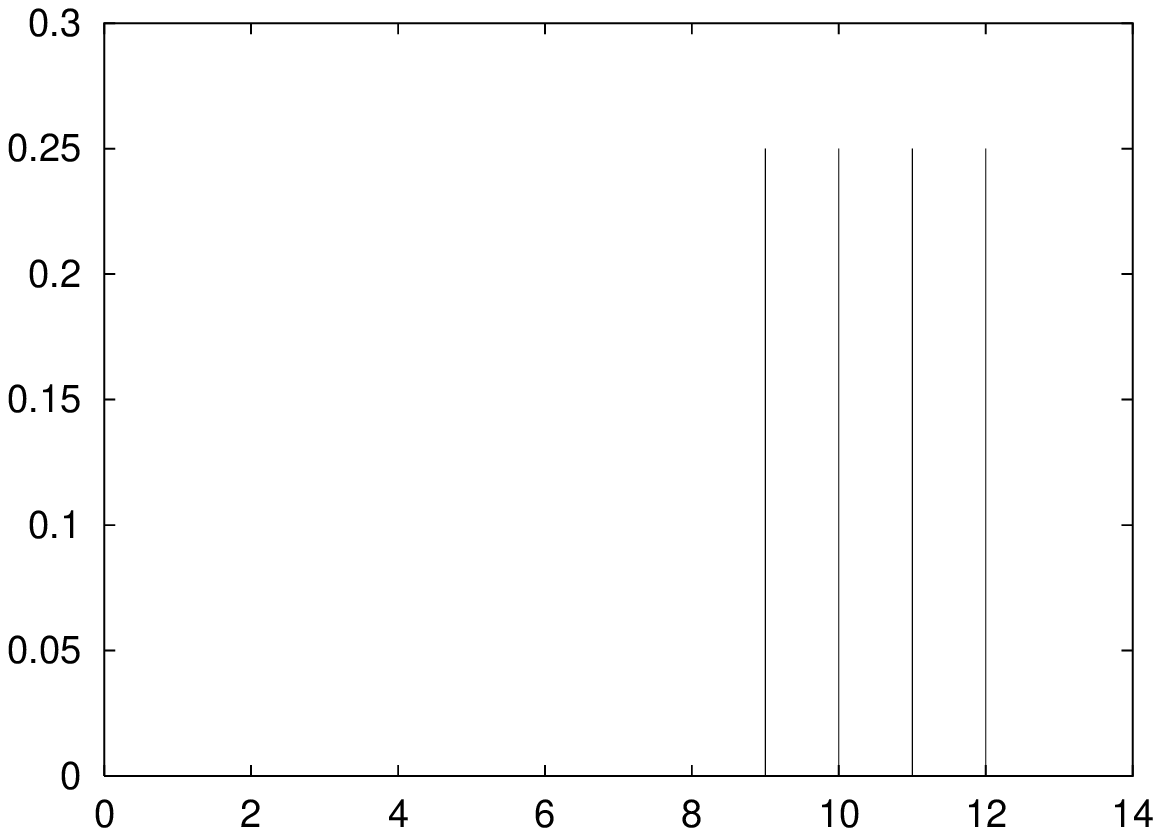,width=110pt,height=100pt}
\\
($D_{3,4}$) 
&
($D_{9,4}$) 
\\
\multicolumn{2}{c}{\psfig{figure=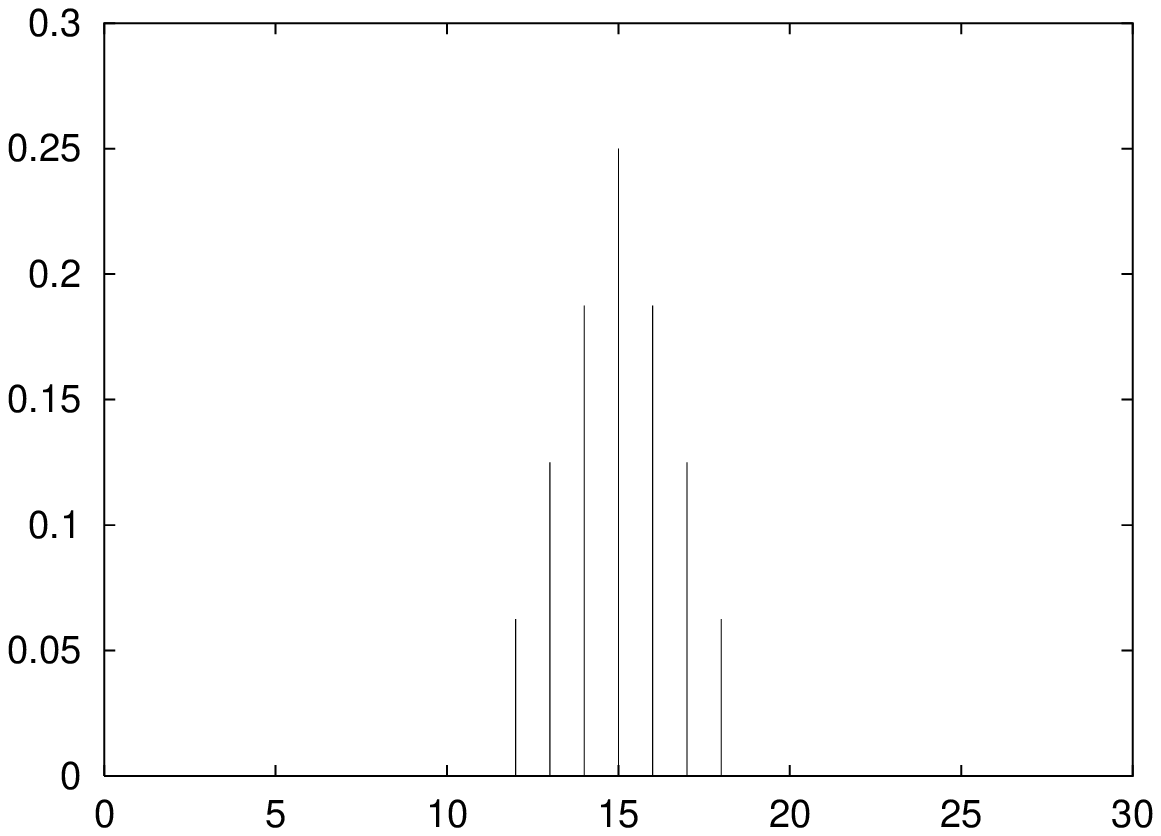,width=200pt,height=100pt}}
\\
\multicolumn{2}{c}{
convolution product ($D_3 \oplus D_9$)}

\end{tabular}
\end{center}
\caption{Example of convolution product}
\label{fig_pc}
\end{figure}

Exhaustive fitness allocation allows only to generate landscapes with small search space (in our experiments $2^{16}$ solutions). Hence we must find a way to construct similar problems on a larger scale.
We propose here to concatenate several small ND-Landscapes to create an additive ND-Landscape.
Even though we are not able to create an additive ND-Landscape directly from a neutral degree distribution we can chose the mean, the standard deviation and less precisely the shape of his neutral degree distribution.
Moreover, this method allows us to know exactly the neutral degree distribution of the resulting landscape.
Let be $P_1 =\ (E_1,V_1,f_1)$ and $P_2 =\ (E_2,V_2,f_2)$ two fitness landscapes.
We define the extended landscape $P = P_1 \oplus P_2 = \ (E,V,f)$ such as :
\begin{itemize}
\item{$E = E_1 \times E_2$}
\item{$\forall{(x_1,y_1) \in E_1^2},\forall{(x_2,y_2) \in E_2^2},$\\\ \ \ \ $\ (x_1,x_2)\ \in V(y_1,y_2) \iff (x_1 \in V_1(y_1) \ and\  x_2 = y_2) \ or\ ( x_1 = y_1   \ and\ x_2 \in V_2(y_2) )$}
\item{$f(x_1,x_2) = f_1(x_1) + f_2(x_2)$}
\end{itemize}
 The size of the larger ND-landscape will be the product of the sizes of the small ones.
The neutral degree distribution of the resulting landscape is the \textit{convolution product} of the two components distributions.
The convolution product of two distributions $D_1$ and $D_2$ is the distribution $D$ (see figure \ref{fig_pc}) such as $$\forall{n \in \mbox{I\hspace{-.15em}N}}, D(n) = \sum_{i=0}^{n}(D_1(i) \times D_2(n-i))$$

We have the following properties : \\

 $average(D) = average(D_1) + average(D_2) $

and $\sigma(D) = \sqrt{\sigma^2(D_1)+\sigma^2(D_2)} $ where $\sigma$ is the standard deviation.\\

The convolution product of two normal (resp. $\chi^2$, Poisson) distributions is a normal (resp. $\chi^2$, Poisson) distribution.

Concatenation is commutative, so it is simple to concatenate a number of small landscapes.
Hence, we can tack together many small ND-Landscapes generated exhaustively to obtain a bigger one with known neutral degree distribution.
Moreover, it has been proven by Jones \cite{jones95b} that the FDC coefficient is unchanged when multiple copies of a problem are concatenated to form a larger problem.

\section*{Conclusion}

This paper presents the family of ND-Landscapes as a model of neutral
fitness landscapes. Most of the academic fitness landscapes found in the literature
deals with neutrality, either as an add-on feature or as an incidental
property. Theoretical studies in molecular evolution, as well as in
optimization, show that the distribution of neutral degree is a key factor
in the evolution of population on a neutral network. Hence to understand and
study the influence of neutrality on evolutionary search process it may be
helpful to design landscapes with a given neutral distribution. We propose
three steps to design such landscapes: first using an algorithm we construct
a landscape whose distribution roughly fits the target one, then we use a
simulated annealing heuristic to bring closer the two distributions, and
finally we affect fitness values to each neutral network. The last step can
be used to tune the landscape difficulty according to other criteria than
neutrality. So we can study interaction between neutrality and another
source of hardness. 
In this paper we use trap functions to affect fitness
values, this allows us to study the interplay between deceptiveness
and neutrality; in particular, experimental results show that neutrality
smoothes the correlation between fitness and distance.
For some problems, introducing neutrality could be benefic as shown in
\cite{Ebner01} by increasing the evolvability.
On the other hand, neutrality could destroy useful information such
as correlation as it is shown in our experiments.
 As these landscapes need an exhaustive enumeration of the search space we propose to concatenate small
ND-Landscapes to scale up over 16 bits. Then, using the convolution product
of distribution, we are able to design large landscapes with known neutral
distribution.


Future work along these lines includes studying the influence of various distributions and different ways to affect fitness value on the dynamics of a population on a neutral network. With ND-trap functions we have focused on the interplay between deceptiveness and neutrality. By replacing trap functions by NK functions\cite{kauffman93}, we might be able to highlight correlation between epistasis and neutrality.

\bibliographystyle{unsrt}

\end{document}